\documentclass{article}

\usepackage{arxiv}

\usepackage[utf8]{inputenc} % allow utf-8 input
\usepackage[T1]{fontenc}    % use 8-bit T1 fonts
\usepackage{hyperref}       % hyperlinks
\usepackage{url}            % simple URL typesetting
\usepackage{booktabs}       % professional-quality tables
\usepackage{amsfonts}       % blackboard math symbols
\usepackage{nicefrac}       % compact symbols for 1/2, etc.
\usepackage{microtype}      % microtypography
\usepackage{lipsum}
\usepackage{graphicx}
\usepackage{amsmath}
\usepackage{amssymb}
\usepackage{amsfonts}
\graphicspath{ {./images/} }

\title{Conformal Prediction for Uncertainty Estimation in Drug-Target Interaction Prediction}

\author{
 Morteza Rakhshaninejad \\
  Department of Data Analysis\\ and Mathematical Modeling\\
  Ghent University\\
  \texttt{Morteza.Rakhshaninejad@UGent.be} \\
   \And
Mira J\"{u}rgens \\
  Department of Data Analysis\\ and Mathematical Modeling\\
  Ghent University\\
  \texttt{Mira.Juergens@UGent.be} \\
  \And
 Nicolas Dewolf \\
  Department of Data Analysis\\ and Mathematical Modeling\\
  Ghent University\\
  \texttt{Nicolas.Dewolf@UGent.be} \\
  \And
 Willem Waegeman \\
  Department of Data Analysis\\ and Mathematical Modeling\\
  Ghent University\\
  \texttt{Willem.Waegeman@UGent.be} \\
}

\begin{document}
\maketitle
\begin{abstract}
Accurate drug–target interaction (DTI) prediction with machine learning models is essential for drug discovery. Such models should also provide a credible representation of their uncertainty, but applying classical marginal conformal prediction (CP) in DTI prediction often overlooks variability across drug and protein subgroups. In this work, we analyze three cluster-conditioned CP methods for DTI prediction, and compare them with marginal and group-conditioned CP. Clusterings are obtained via nonconformity scores, feature similarity, and nearest neighbors, respectively. Experiments on the KIBA dataset using four data-splitting strategies show that nonconformity-based clustering yields the tightest intervals and most reliable subgroup coverage, especially in random and fully unseen drug–protein splits. Group-conditioned CP works well when one entity is familiar, but residual-driven clustering provides robust uncertainty estimates even in sparse or novel scenarios. These results highlight the potential of cluster-based CP for improving DTI prediction under uncertainty.
\end{abstract}

% keywords can be removed
\keywords{Conformal Prediction\and Drug–target Interaction\and Uncertainty Quantification\and Cluster-conditioned Prediction\and Prediction Intervals}

\section{Introduction}
\label{sec:intro}
Drug–target interaction (DTI) prediction is a central task in computational drug discovery, enabling the identification of candidate compounds likely to bind to specific protein targets~\cite{ozturk2018deepdta}. Machine learning models have significantly improved the scalability and accuracy of DTI screening by using chemical and biological features to predict binding affinities. However, most existing approaches provide only point estimates of interaction strength, without quantifying the uncertainty of their predictions. This lack of uncertainty estimation can mislead downstream validation, undermining the reliability of experimental decisions. 
To address this, uncertainty quantification (UQ) methods have emerged as crucial tools, particularly those that offer predictive intervals for regression and prediction sets for classification alongside point estimates. Among these, conformal prediction (CP) provides a distribution-free framework for generating calibrated prediction intervals with finite-sample coverage guarantees~\cite{vovk2005algorithmic,angelopoulos2021gentle}. In regression tasks, CP methods wrap around any predictive model to output prediction intervals that contain the true label with a user-specified probability \(1 - \alpha\), under the assumption of data exchangeability. 
While marginal CP ensures coverage on average across all test points, it can fail to capture heterogeneous error behavior across subpopulations, such as distinct drug or protein families. Specifically, marginal coverage satisfies $P(y \in \widehat{C}(\mathbf{x})) \ge 1 - \alpha$, where the probability is taken over the joint distribution of test inputs and labels. However, this guarantee does not imply that all subgroups receive adequate coverage. In contrast, conditional coverage aims to satisfy $P(y \in \widehat{C}(\mathbf{x}) \mid \mathbf{x}) \ge 1 - \alpha$ for every input $\mathbf{x}$, a much stronger criterion that is generally impossible to achieve without further assumptions. Therefore, to improve practical utility in the presence of data heterogeneity, we investigate methods that approximate conditional coverage by conditioning on meaningful subgroups.

In this study, we investigate the application of a cluster-conditioned conformal prediction (CCP) framework to provide more accurate and personalized uncertainty estimates in DTI regression. Building on recent advances in class-conditional CP~\cite{ding2023class}, we propose two clustering-based variants—one based on nonconformity score distributions (CCP-NC) and the other on clustering input features (CCP-FC), as well as a nearest-neighbor--based variant (CCP-NN). A key novelty of our approach is that we construct calibration sets by conditioning on both components of a drug–target interaction, rather than treating the interaction as a single atomic unit, allowing prediction intervals to more accurately reflect uncertainty arising from either side. We benchmark these CCP methods against standard Marginal CP (MCP)~\cite{vovk2005algorithmic} and Group-Conditioned CP (GCP)~\cite{vovk2003mondrian}, where drug and protein identities are used as group labels.
Our experiments use the KIBA dataset~\cite{tang2014making}, with a gradient boosting regressor and feature representations combining molecular fingerprints, chemical descriptors, amino acid composition, and transformer-based embeddings. To evaluate robustness and generalization, we implement four data-splitting strategies that progressively increase the dissimilarity between training and test sets.
We introduce CCP framework for DTI regression, comprising three variants: CCP-NC, CCP-FC, and CCP-NN. These models are benchmarked against MCP and GCP across all four data-splitting strategies, including the challenging New Drug–Protein split. To improve coverage at the subgroup level, we propose a joint configuration strategy that tunes both the number of clusters and the calibration set ratio. Finally, we evaluate all methods using global coverage, prediction interval width, and subgroup-level coverage gap.

The remainder of the paper is organized as follows: Section~\ref{sec:related-work} reviews prior work on CP and its applications in computational biology. Section~\ref{sec:sec2} describes the dataset, predictive model, data-splitting strategies, and CP methodologies used in this study. Section~\ref{sec:experiments} presents experimental results comparing different CP variants across various evaluation criteria. Section~\ref{sec:conclusion} concludes the paper and outlines directions for future work.

\section{Related Work}
\label{sec:related-work}
CP estimates a threshold on conformity scores to construct prediction intervals or set-valued predictions, ensuring that the error rate does not exceed a user-specified level $\alpha$ in finite samples \cite{vovk2005algorithmic}. In the original transductive scheme, the underlying predictive model is re-trained for every candidate label of the test point \cite{Saunders1999Trans}. The more practical split/inductive variant \cite{papadopoulos2002inductive,angelopoulos2021gentle} instead trains a single predictive model on a proper-training set and reserves a disjoint calibration set to estimate the $(1-\alpha)$-quantile of a nonconformity score. The choice of NCS, which quantifies how atypical an observation is, controls the efficiency of the resulting intervals and has spurred extensive research on score design, normalization, and learning \cite{kato2023review,bellotti2020constructing}.
Split-conformal guarantees hold only marginally, i.e.\ on average over the population; particular subgroups may be under- or over-covered. Mondrian, or label-conditional, CP addresses this by calibrating within each predefined group to achieve exact conditional coverage~\cite{vovk2003mondrian}, but the per-group quantile becomes unstable when few calibration samples are available, leading to very wide intervals. Recent theoretical work also shows that, for heteroskedastic data, normalizing nonconformity scores by estimated noise levels can achieve approximate conditional validity without requiring explicit group splits~\cite{dewolf2023conditional}.

Classical CP methods ensure marginal validity but do not guarantee conditional validity, a stricter and often more desirable property that tailors prediction regions to individual covariates. Prior work has shown that constructing non-trivial prediction sets with exact conditional validity from finite samples is impossible without making strong assumptions on the data distribution~\cite{vovk2012Conditional,Lei2013distribution,foygel2021limits}. Consequently, most research focuses on achieving approximate conditional validity while retaining distribution-free guarantees~\cite{gibbs2025conformal}. A common relaxation involves group-conditional guarantees~\cite{ding2023class,jung2022batch}, where coverage is ensured for a predefined set of groups. Another line of work addresses covariate shift, relaxing the exchangeability assumption, and imposes conditional guarantees with respect to a fixed family of shifts~\cite{tibshirani2019conformal}. However, this is distinct from conditional conformal prediction under exchangeability, and may miss shifts tied to the predictive model's uncertainty. Efforts to broaden the covariate shift space~\cite{gibbs2025conformal} risk conservatism and unnecessarily wide intervals, as noted by Hore and Barber~\cite{hore2023conformal}. Alternative approaches, such as localised CP~\cite{guan2023localized}, weigh the residuals on the calibration set by a similarity function between test and calibration samples, although such methods can struggle in high-dimensional feature spaces. Recent clustered variants merge groups with similar score distributions before calibration, balancing validity and efficiency, and have proved effective in image recognition and property-valuation tasks~\cite{ding2023class,hjort2024clustered,lim2021normalized}. Other work pursues softer relaxations such as local or normalized coverage~\cite{angelopoulos2021gentle}. Together these advances shift the focus from global to equitable, subgroup-aware uncertainty quantification.

In computational biology, CP has also seen emerging applicationss. Boger et al.~\cite{boger2025functional} applied CP principles to protein structure and homology detection, enabling robust pre-filtering for structural alignment and achieving state-of-the-art performance in enzyme classification without retraining predictive models. In drug discovery, Laghuvarapu et al.~\cite{laghuvarapu2023codrug} proposed CP-based prediction sets for molecular properties under covariate shift, significantly reducing the coverage gap, by over 35\%, compared to standard CP methods. Within DTI prediction, most studies focus on point estimates of binding affinity~\cite{ozturk2018deepdta,iliadis2024comparison}, while CP remains underexplored. A notable exception is Orsolic and Smuc~\cite{orvsolic2023dynamic}, who proposed a local conformal approach using dynamically constructed calibration sets based on neighborhood similarity. However, their method does not leverage cluster-based conformal predictors, which our experiments show to be more effective in controlling the subgroup-level coverage gap. Thus, the application of CP to DTI remains largely unexplored, particularly in terms of generating calibrated prediction intervals with conditional coverage guarantees at the level of individual drugs and proteins, a gap this work aims to address.

\section{Materials and Methods}\label{sec:sec2}
\subsection{Dataset and Feature Representations}\label{sec:dataset}
We constructed our dataset by integrating multiple feature modalities for drugs and protein targets to capture the chemical and biological variability of kinase-inhibitor interactions. Specifically, SMILES strings \cite{weininger1988smiles} were used to represent small-molecule inhibitors \( d_i \), and primary amino acid sequences were used to represent protein targets \( t_j \). Each drug-target interaction is defined as a unique pair \( (d_i, t_j) \).
Binding affinities (\( K_d \)) were transformed into dimensionless values using \( y(d_i, t_j) = -\log_{10}(K_d / 10^9) \), where \( y(d_i, t_j) \) denotes the transformed affinity between drug \( d_i \) and target \( t_j \).
We evaluated our models exclusively on the KIBA benchmark dataset \cite{tang2014making}, which comprises 2068 small-molecule inhibitors, 229 protein kinase targets, and 118,036 recorded drug–target interactions.

\begin{figure}[t]
	\centering % Often used to center the figure
	\includegraphics[width=0.7\linewidth]{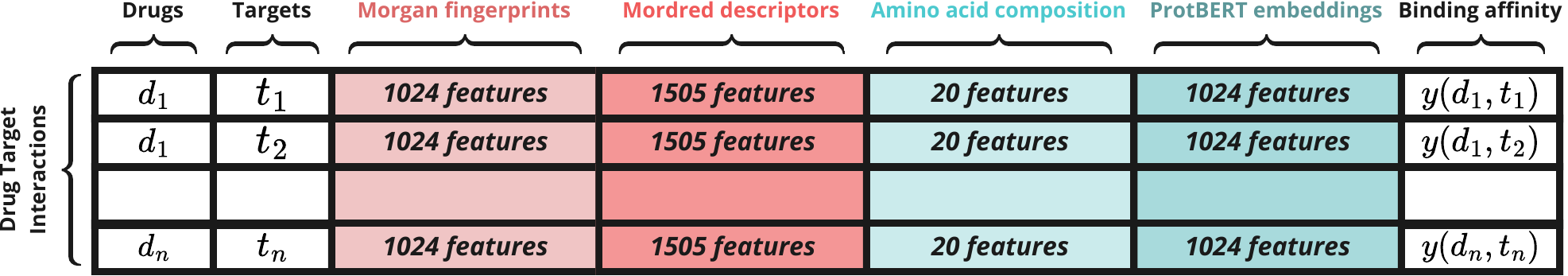}
	\caption{Schematic of dataset structure: each interaction is a unique drug--target pair \( (d_i, t_j) \) with binding affinity \( y(d_i, t_j) \).}
	\label{fig:dataset} % Label should typically go after the caption
\end{figure}

Each drug was represented using Morgan fingerprints with 1024 features, which encode molecular substructures using circular fingerprints with a radius of 2~\cite{rogers2010extended}, and Mordred descriptors, comprising 1505 features that capture a comprehensive set of physicochemical, topological, and constitutional molecular properties~\cite{moriwaki2018mordred}. Protein targets were encoded using Amino Acid Composition, a 20-feature representation that quantifies the frequency of each amino acid within the sequence, and ProtBERT embeddings, which consist of 1024-dimensional transformer-based representations trained on protein sequences~\cite{elnaggar2021prottrans}.
Motivated by their complementary strengths, we combined four feature types, Morgan fingerprints, Mordred descriptors, Amino Acid Composition, and ProtBERT embeddings, to capture diverse chemical and biological properties at both local and global levels. The resulting drug--target representations were concatenated into a feature matrix \( X \in \mathbb{R}^{N \times 3573} \), where \( N \) is the number of interactions.
Figure~\ref{fig:dataset} illustrates the structured representation of our dataset, showing the feature modalities used for each pair of interactions.

\subsection{Predictive Model}\label{sec:predModel}
To estimate the binding affinity of the drug-target, we used the gradient boosting regressor \cite{friedman2001greedy}, a widely used ensemble learning technique recognized for its efficiency, scalability, and ability to model complex non-linear relationships in tabular data. Its integrated feature selection and regularization mechanisms make it particularly suitable for high-dimensional tasks such as DTI prediction. The trained model maps each input feature vector \(\mathbf{x}(d_i, t_j) \in \mathbb{R}^{N \times 3573}\), representing a drug--target pair, to a continuous output \(\hat{y}(d_i, t_j) = f_{\theta}(\mathbf{x}(d_i, t_j))\), where \(f_{\theta}\) denotes the gradient boosting regressor model and \(\theta\) represents the learned parameters. We configured the model with squared error as the loss function, 500 boosting stages, a learning rate of 0.05, and a maximum tree depth of 6. The implementation was carried out using the \texttt{GradientBoostingRegressor} class from the \texttt{scikit-learn} Python library (version 1.3.0).

\subsection{Dataset Splitting Strategies}\label{sec:split_strategies}
To evaluate the robustness and generalization capability of our model, we employed four dataset splitting strategies. Each strategy ensures that the model is evaluated under different distributional constraints, ranging from favorable overlap to strict disjointness. In every case, the dataset is partitioned into three disjoint subsets: a training set ($\mathcal{D}_{\text{train}}$) for model fitting, a calibration set ($\mathcal{D}_{\text{cal}}$) for CP, and a test set ($\mathcal{D}_{\text{test}}$) for final evaluation. 
All strategies maintain a 50\%–25\%–25\% split across training, calibration, and test data. The four splitting strategies are defined as follows. In the Random Split, interaction-level random splitting is performed: 50\% of all interaction samples are assigned to the training set, and the remaining 50\% are then split equally into calibration and test sets. Since calibration and test samples are drawn from the same pool, they are exchangeable by construction. This split imposes the least restriction and assumes that drugs and proteins are evenly distributed across all subsets.
The Drug Split and Protein Split strategies follow the same mechanism. In the Drug Split, 50\% of unique drugs are randomly selected, and all their corresponding interactions are assigned to the training set. The remaining 50\% of drugs are randomly divided into two equal groups to form the calibration and test sets. Similarly, in the Protein Split, 50\% of unique proteins are used for training, and the remaining proteins are split equally into calibration and test sets. In both cases, interactions are assigned based on entity membership, ensuring that no drug or protein appears in more than one subset, while the other side of the interaction (proteins or drugs) may be shared. Calibration and test sets are constructed from a common pool of interactions involving unseen entities, preserving exchangeability.
Finally, the New Drug–Protein Split represents the most challenging scenario. Both the drugs and proteins in the calibration and test sets are completely unseen during training. Specifically, 50\% of drugs and 50\% of proteins are randomly selected to form the training set, which includes only interactions where both entities belong to the training pool. The remaining interactions—comprising combinations of unseen drugs and proteins—form the held-out pool, which is then equally divided into calibration and test sets. This split ensures exchangeability but may result in a reduced number of available samples due to its stringent filtering criteria.

\begin{figure}[t]
	\centering % Often used to center the figure
	\includegraphics[width=0.75\linewidth]{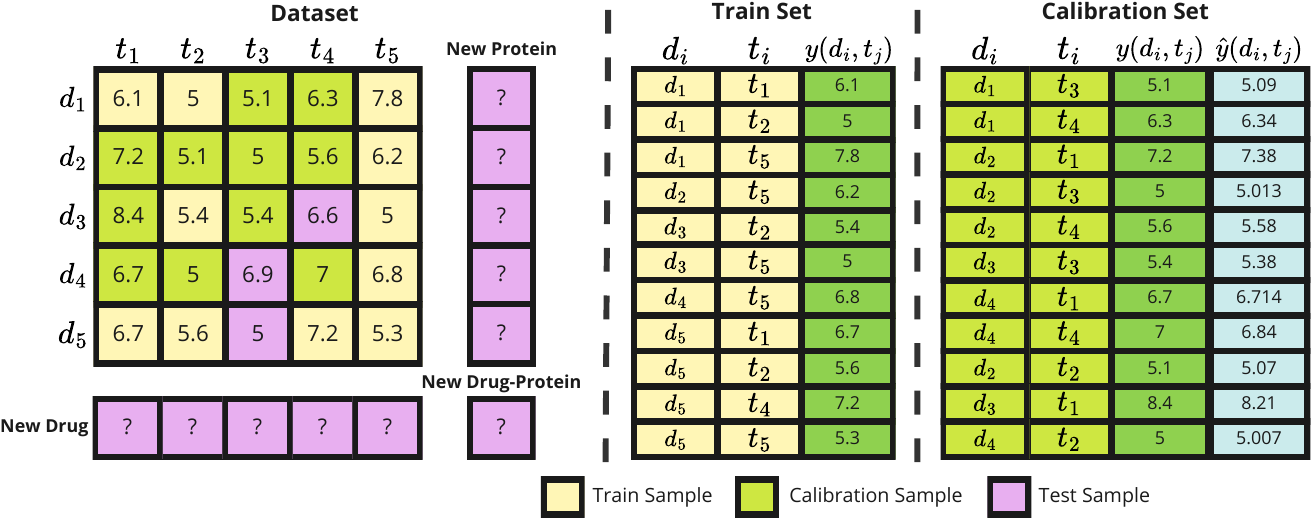}
	\caption{Illustrative example of dataset preparation for model training and conformal prediction. The drug--target interaction matrix shows samples used for training, calibration, and testing.}
	\label{fig:splitting_strategies} % Label should typically go after the caption
\end{figure}

A visual illustration of the dataset splitting strategies is shown in Figure~\ref{fig:splitting_strategies}. The left panel represents the drug–target interaction matrix with binding affinity values. The center panel highlights the subset of interactions used to train the predictive model, while the right panel shows the calibration set samples, whose affinities are predicted using the trained model and used for conformal quantile estimation. Each split strategy defines how these sets are constructed and reflects a different level of generalization difficulty, from the permissive Random Split to the stringent New Drug–Protein Split.

\subsection{Problem statement}\label{sec:cp_general}
Let \(\mathcal{D} = \{(\mathbf{x}_i, y_i)\}_{i=1}^{N}\) be a labeled dataset of \(N\) samples, where each \(\mathbf{x}_i \in \mathbb{R}^{N \times 3573}\) is a feature vector representing an interaction between drug \( d_i \) and target \( t_j \), and \( y_i \in \mathbb{R} \) is the corresponding transformed binding affinity label. We consider a point prediction function \(f_\theta: \mathbb{R}^{N \times 3573} \to \mathbb{R}\), which in our case is realized by the \texttt{GradientBoostingRegressor} model described in Section~\ref{sec:predModel}. Our goal is to construct a predictive interval \(\widehat{C}(\mathbf{x})\) for a new test input \(\mathbf{x}\), such that the probability of \(\widehat{C}(\mathbf{x})\) containing the true value \(y\) is approximately \(1-\alpha\). We typically partition the dataset of \(N\) samples into three disjoint subsets 
\(\mathcal{D}_{\text{train}}, \mathcal{D}_{\text{cal}}, \mathcal{D}_{\text{test}}\) (as we discussed in Section~\ref{sec:split_strategies})
of respective sizes \(n_{\text{train}}, n_{\text{cal}}, n_{\text{test}}\), with $n_{\text{train}} + n_{\text{cal}} + n_{\text{test}} \;=\; N$. Denoting the samples by $(\mathbf{x}_i, y_i)$, we train the model 
\(f_\theta\) on \(\mathcal{D}_{\text{train}}\). The calibration set \(\mathcal{D}_{\text{cal}}\) is used to compute nonconformity scores and estimate the relevant quantile, which determines how wide the prediction interval must be to ensure the desired coverage level $1-\alpha$. Finally, \(\mathcal{D}_{\text{test}}\) is reserved for evaluation.

A key component of CP is the choice of a nonconformity (or conformity) measure, which quantifies how unusual an observed label \(y_i\) is relative to the model’s prediction \(f_\theta(\mathbf{x}_i)\). A common choice in regression is the absolute residual~\cite{papadopoulos2011regression}:
\begin{equation}\label{eq:absolute_residual}
    s_i = |y_i - f_\theta(\mathbf{x}_i)|.
\end{equation}
However, if the data is heteroscedastic (i.e., the noise level varies across the input space), the efficiency of prediction intervals may degrade. In such cases, a normalized nonconformity score can be used~\cite{papadopoulos2002inductive}:
\begin{equation}\label{eq:normalized_nonconformity}
    s_i = \frac{|y_i - f_\theta(\mathbf{x}_i)|}{\sigma_\theta(\mathbf{x}_i)},
\end{equation}
where \(\sigma_\theta(\mathbf{x}_i)\) is an estimate of the local uncertainty for \(\mathbf{x}_i\). In our case, since we applied a Box--Cox transformation \cite{box1964analysis} to normalize the input binding affinity values, we assume that the transformed data exhibit homoscedastic behavior. Consequently, we adopt the absolute residual form in Equation~\eqref{eq:absolute_residual} as our nonconformity score throughout this study.
After computing the nonconformity scores \(s_i\) on the calibration set, we sort them in non-decreasing order: $s_{(1)} \le s_{(2)} \le \dots \le s_{(n_{\text{cal}})}$, and select the conformal quantile threshold:
\begin{equation}\label{eq:quantile_calculation}
    \hat{q} \;=\; s_{(\lceil (1-\alpha)\,(n_{\text{cal}}+1)\rceil)},
\end{equation}
which corresponds to the \((1-\alpha)\)-quantile of the empirical distribution of the nonconformity scores \(s_i\). 
The construction of the prediction interval \(\widehat{C}(\mathbf{x})\) depends on the choice of the nonconformity function. If the absolute residual is used as the nonconformity score (see Equation~\ref{eq:absolute_residual}), the prediction interval is defined as
\begin{equation}\label{eq:cp_interval_absolute}
    \widehat{C}(\mathbf{x}) \;=\; \left[\, f_\theta(\mathbf{x}) - \hat{q},\;\; f_\theta(\mathbf{x}) + \hat{q} \,\right].
\end{equation}
Alternatively, if the normalized residual is used (see Equation~\ref{eq:normalized_nonconformity}), incorporating an estimate of local uncertainty \(\sigma_\theta(\mathbf{x})\), the prediction interval becomes
\begin{equation}\label{eq:cp_interval_normalized}
    \widehat{C}(\mathbf{x}) \;=\; \left[\, f_\theta(\mathbf{x}) - \hat{q} \cdot \sigma_\theta(\mathbf{x}),\;\; f_\theta(\mathbf{x}) + \hat{q} \cdot \sigma_\theta(\mathbf{x}) \,\right].
\end{equation}

Under appropriate conditions—most notably, exchangeability between the calibration and test samples—this interval satisfies the approximate coverage guarantee \( P\bigl( y \in \widehat{C}(\mathbf{x})\bigr) \approx 1 - \alpha \). In what follows, we describe CP approaches. These methods differ in how the calibration set is partitioned (if at all) and whether the coverage guarantee is interpreted globally or conditionally (e.g., within a group or cluster).

\subsubsection{Marginal Conformal Prediction}
\label{sec:marginal_cp}
Marginal Conformal Prediction (MCP) does not condition on any subgroup or feature partition \cite{papadopoulos2002inductive}. It uses a single global calibration set to compute prediction intervals. This approach represents the simplest form of CP, as it assumes no additional structure in the data beyond the exchangeability assumption.
We partition the dataset \(\mathcal{D}\) into $\mathcal{D}_{\text{train}}$, $\mathcal{D}_{\text{cal}}$, and $\mathcal{D}_{\text{test}}$, and train the regression model \(f_\theta\) on $\mathcal{D}_{\text{train}}$ to predict binding affinity values \( y(d_i, t_j) \) from the concatenated feature vector \( \mathbf{x}(d_i, t_j) \).
This training process is applied separately for each dataset splitting strategy described in Section~\ref{sec:split_strategies}.
Next, we compute the nonconformity scores for each calibration sample \((\mathbf{x}_i, y_i) \in \mathcal{D}_{\text{cal}}\), using the absolute residual defined in Equation~\ref{eq:absolute_residual}.
The scores are then sorted in non-decreasing order, and the conformal threshold \(\hat{q}\) is selected as the \((1-\alpha)\)-quantile.

For any test sample $(\mathbf{x}_i, y_i) \in \mathcal{D}_{\text{test}}$, where $\mathbf{x}_i \in \mathcal{X}$, the model produces a point prediction $\hat{y}_i = f_\theta(\mathbf{x}_i)$ and constructs a prediction interval, as defined in Equation~\ref{eq:cp_interval_absolute}. Under the exchangeability assumption, this procedure guarantees that the true label $y_i$ will fall within the predictive interval $\widehat{C}(\mathbf{x}_i)$ with probability at least $1 - \alpha$, i.e., $P(y \in \widehat{C}(\mathbf{x})) \ge 1 - \alpha.$
This approach provides marginal coverage across all test samples without conditioning on groups or structure. While simple and broadly applicable, it produces uniformly wide intervals (unless scaled by input variance), making it insensitive to underlying data heterogeneity.

\subsubsection{Group-Conditioned Conformal Prediction}
\label{sec:group_cp}
Group-Conditioned Conformal Prediction (GCP) is based on the concept of Mondrian CP \cite{vovk2005algorithmic}, which extends the standard marginal approach by conditioning on known group structures in the data. The goal is to obtain valid coverage within each group, rather than across all test instances. Let \(g(\mathbf{x})\) denote a group assignment function that maps each input \(\mathbf{x} \in \mathcal{X}\) to one of \(G\) distinct groups, indexed by \(g \in \{1, \dots, G\}\). After training the model \(f_\theta\) on \(\mathcal{D}_{\text{train}}\), we calculate nonconformity scores separately within each group of the calibration set.
In contrast to MCP, which applies a single global threshold, GCP divides the calibration samples into finer-grained categories. In our DTI application, natural groupings emerge on the basis of drug and protein identities. Specifically, we consider each drug and each protein as defining a Mondrian category, allowing the prediction intervals to adapt to the uncertainty associated with individual compounds or targets. This group-based structure is especially important in data-splitting strategies such as drug split, protein split, or new drug–protein split (see Section~\ref{sec:split_strategies}), where the challenge of generalization varies according to which molecular entities are held.

For each calibration sample \((\mathbf{x}_i, y_i) \in \mathcal{D}_{\text{cal}}\), we compute a nonconformity score \(s_i\). These scores are grouped using a group function \(g(\cdot)\), which can correspond to the drug identity, protein identity, or both.
For each group \(g\), we collect the corresponding nonconformity scores and sort them: $s_{(1)}^g \le s_{(2)}^g \le \dots \le s_{(n_g)}^g$, where \(n_g\) is the number of calibration samples assigned to group \(g\). The group-specific quantile is then computed as $\hat{q}_g = s_{(\lceil (1-\alpha)(n_g + 1) \rceil)}^g$.
At test time, for a given interaction \((d_{\text{test}}, t_{\text{test}})\), we determine whether the test drug or test protein is represented in the calibration set. 
The quantile threshold \(\hat{q}_{g^*}\) is selected based on the availability of the test drug and protein in the calibration set. If both the test drug and test protein appear in the calibration set, their corresponding groups are merged, and a combined quantile is computed from the union of their nonconformity scores. If only the test drug is present, the quantile threshold is taken from its respective group. Similarly, if only the test protein appears, the quantile is based on its group. If neither the drug nor the protein is observed in the calibration set, a global quantile computed from all calibration scores is used instead.
This logic is formalized as:
\begin{equation}\label{eq:gcp_combined_quantile}
    \hat{q}_{g^*} =
    \begin{cases}
        \hat{q}(\{s_i : d_i = d_{\mathrm{test}} \lor t_i = t_{\mathrm{test}}\}, 1-\alpha), & \text{if } d_{\mathrm{test}}, t_{\mathrm{test}} \in \mathcal{D}_{\mathrm{cal}},\\[1ex]
        \hat{q}(\{s_i : d_i = d_{\mathrm{test}}\}, 1-\alpha), & \text{if } d_{\mathrm{test}} \in \mathcal{D}_{\mathrm{cal}},\; t_{\mathrm{test}} \notin \mathcal{D}_{\mathrm{cal}},\\[1ex]
        \hat{q}(\{s_i : t_i = t_{\mathrm{test}}\}, 1-\alpha), & \text{if } t_{\mathrm{test}} \in \mathcal{D}_{\mathrm{cal}},\; d_{\mathrm{test}} \notin \mathcal{D}_{\mathrm{cal}},\\[1ex]
        \hat{q}_{\mathrm{global}}, & \text{otherwise}.
    \end{cases}
\end{equation}
The final prediction interval for test input \(\mathbf{x}\) is given by~$\widehat{C}(\mathbf{x}) = \left[\, f_\theta(\mathbf{x}) - \hat{q}_{g^*},\; f_\theta(\mathbf{x}) + \hat{q}_{g^*} \,\right]$.
Assuming exchangeability holds within each group, this procedure yields the following group-conditional coverage guarantee~\cite{vovk2003mondrian}: \(P\left( y \in \widehat{C}(\mathbf{x}) \mid g(\mathbf{x}) = g \right) \ge 1 - \alpha\), for all \(g \in \{1, \dots, G\}\).

Compared to MCP, this method provides more fine-grained, context-aware intervals that adapt to variability between groups. However, its effectiveness depends on having a sufficient number of calibration samples in each group to ensure stable quantile estimation \cite{ding2023class}. In the DTI setting, this is especially important when generalizing to novel drugs or targets not seen in training, as group-conditioned thresholds allow the conformal predictor to keep coverage within these molecular categories.

\subsubsection{Cluster-Conditioned Conformal Prediction: Nonconformity-Based Clustering}
\label{sec:ccp_nc}
In some applications, explicit grouping variables may be unavailable, or the number of samples within each group may be too small to reliably estimate group-specific quantiles. In such cases, we can partition the calibration set in an unsupervised manner using clustering, which allows us to identify latent subpopulations that may exhibit different noise or uncertainty characteristics. This gives rise to CCP, which leverages data-driven clusters to construct calibration sets and potentially improve local coverage properties.
In the Nonconformity-Based Clustering variant (CCP-NC), clustering is performed directly on the nonconformity scores of the calibration samples \cite{ding2023class}. After training the predictive model \(f_\theta\) on \(\mathcal{D}_{\text{train}}\), we compute nonconformity scores for each calibration sample \((\mathbf{x}_i, y_i)\in\mathcal{D}_{\text{cal}}\). These scores are then used as input to a clustering algorithm, \(k\)-means, which assigns each sample to one of \(K\) clusters via a mapping \(\kappa: \mathcal{X} \to \{1,\dots,K\}\).
To enable adaptive thresholding, the calibration set \(\mathcal{D}_{\text{cal}}\) is partitioned into two disjoint subsets \cite{hjort2024clustered}. The first subset is used to perform clustering, grouping drugs and proteins based on the distribution of their nonconformity scores. The second subset is then used to compute cluster-specific quantiles, which are used for constructing the final prediction intervals.
The splitting ratio is controlled by a parameter \(\gamma \in \{0.25, 0.5, 0.75\}\), with higher \(\gamma\) values allocating more data to clustering and lower values favoring quantile estimation.

\begin{figure}[t]
	\centering % Often used to center the figure
	\includegraphics[width=0.85\linewidth]{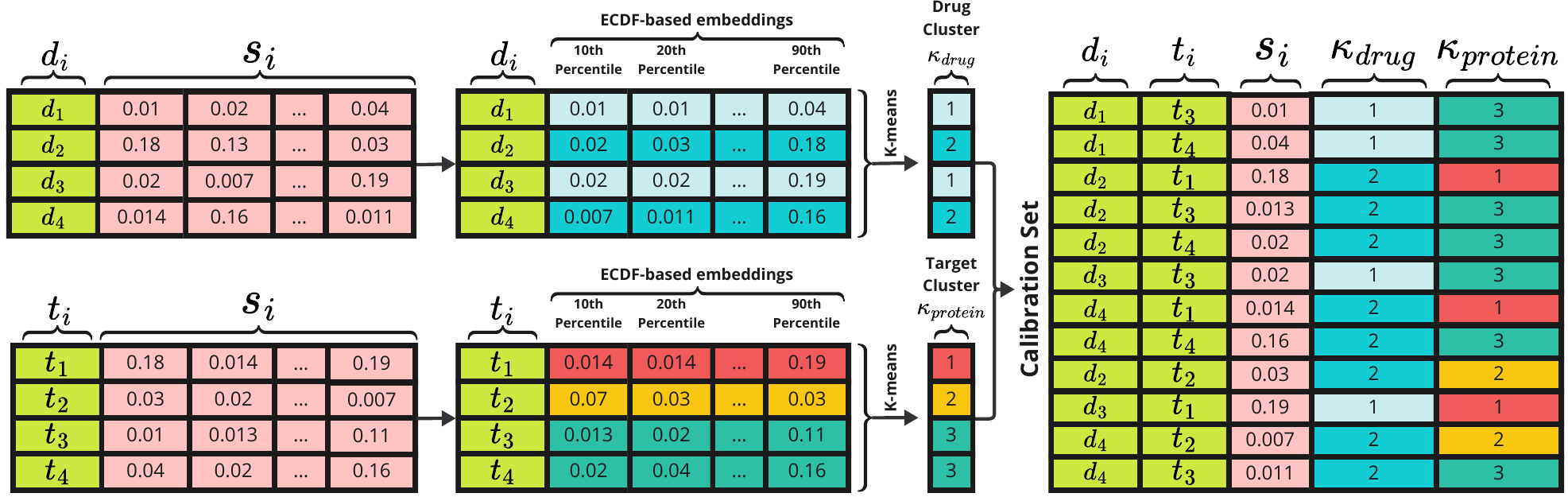}
	\caption{Two-level clustering in CCP-NC. \textbf{Left:} Nonconformity scores grouped by drug and protein. \textbf{Middle:} ECDF embeddings (10th--90th percentiles) extracted. \textbf{Right:} \(k\)-means assigns cluster indices \(\kappa_{\text{drug}}\), \(\kappa_{\text{protein}}\) for cluster-specific quantile computation.}
	\label{fig:clustering} % Label should typically go after the caption
\end{figure}

We then perform two distinct clustering operations over the residuals in the calibration set. First, for each drug \(d_i\), we construct an empirical cumulative distribution function (ECDF) over its nonconformity scores and extract the 10th to 90th percentiles to form a fixed-length embedding vector. Using these embeddings, we apply \(k\)-means clustering with Euclidean distance to assign each drug to a unique drug cluster. Similarly, for each protein \(t_j\), we extract percentile-based ECDF embeddings from its nonconformity scores and apply a second \(k\)-means clustering to assign each protein to a unique protein cluster. As a result, each calibration sample \((d_i, t_j)\) inherits a pair of cluster memberships \((\kappa_{\text{drug}}(d_i), \kappa_{\text{protein}}(t_j))\). This process is illustrated in Figure~\ref{fig:clustering}, where nonconformity score matrices are transformed into percentile-based feature vectors for clustering, and the resulting cluster indices are appended to the calibration set.
The number of clusters \(n_{\text{clusters}}\) and the calibration set splitting ratio \(\gamma\) are selected jointly to optimize subgroup coverage gaps. As introduced in Section~\ref{sec:evaluation_metrics}, the Mean Absolute Coverage Gap (MACG) quantifies the deviation from the target coverage level for drugs and proteins. To identify the best configuration, we evaluate all combinations of \(\gamma \in \{0.25, 0.5, 0.75\}\) and \(\,n_{\mathrm{clusters}}\in\{1,\,5,\,10,\,15,\,20,\,25,\,30,\,35,\,40,\,45,\,50\}\), resulting in 33 possible settings. The optimal pair \((\gamma, n_{\text{clusters}})\) is selected by minimizing the average MACG across drug and protein subgroups:
\begin{equation}\label{eq:combine_MACG}
    \min_{\gamma \in \{0.25, 0.5, 0.75\}, \; n_{\mathrm{clusters}} \in [1,50]} \left( \frac{\text{MACG}_{\mathrm{drug}}(\gamma, n_{\mathrm{clusters}}) + \text{MACG}_{\mathrm{protein}}(\gamma, n_{\mathrm{clusters}})}{2} \right)
\end{equation}
This strategy enables an optimal trade-off between reliability and subgroup-level validity, yielding accurate prediction intervals across heterogeneous biological subgroups.

Using the second portion of the calibration set, we compute cluster-specific nonconformity thresholds for each drug and protein cluster. For a test interaction \((d_i, t_j)\), we denote the cluster assignments by \(\kappa^*(d_i)\) and \(\kappa^*(t_j)\), respectively. The corresponding quantile thresholds are computed from the nonconformity scores \(s_k\) of calibration samples as follows:
\begin{align}
    \hat{q}_{\kappa^*(d_i)} &= \hat{q}(\{s_k : \kappa_{\text{drug}}(d_k) = \kappa^*(d_i)\}, 1 - \alpha), \label{eq:ccp_quantile_drug} \\
    \hat{q}_{\kappa^*(t_j)} &= \hat{q}(\{s_k : \kappa_{\text{protein}}(t_k) = \kappa^*(t_j)\}, 1 - \alpha). \label{eq:ccp_quantile_prot}
\end{align}
If a drug or protein in the test set belongs to a cluster not represented in the quantile estimation subset, we default to the global threshold \(\hat{q}_{\mathrm{global}}\). The final quantile threshold \(\hat{q}_{\mathrm{c}}\) is selected based on the availability of cluster-level information:
\begin{equation}\label{eq:ccp_combined_quantile}
    \hat{q}_{\mathrm{c}} =
    \begin{cases}
        \hat{q}(\{s_k : \kappa_{\text{drug}}(d_k) = \kappa^*(d_i) \lor \kappa_{\text{protein}}(t_k) = \kappa^*(t_j)\}, 1 - \alpha), & \text{if } d_{\mathrm{i}}\text{ and } t_{\mathrm{j}} \in \mathcal{D}_{\mathrm{cal}},\\[1ex]
        \hat{q}_{\kappa^*(d_i)}, & \text{if only } d_{\mathrm{i}} \in \mathcal{D}_{\mathrm{cal}},\\[1ex]
        \hat{q}_{\kappa^*(t_j)}, & \text{if only } t_{\mathrm{j}} \in \mathcal{D}_{\mathrm{cal}},\\[1ex]
        \hat{q}_{\mathrm{global}}, & \text{otherwise}.
    \end{cases}
\end{equation}
This logic ensures that the prediction interval for a test input \(\mathbf{x} = (d_i, t_j)\) is informed by the most relevant cluster-level statistics available. If both clusters are represented in the calibration data, we pool their nonconformity scores to compute a combined threshold. Otherwise, we rely on the available side, or default to the global threshold when necessary. The prediction interval is then formed as:
\begin{equation}\label{eq:ccp_interval}
    \widehat{C}(\mathbf{x}) = \left[ f_\theta(\mathbf{x}) - \hat{q}_{\mathrm{c}},\; f_\theta(\mathbf{x}) + \hat{q}_{\mathrm{c}} \right].
\end{equation}
Provided the nonconformity scores within each cluster are approximately exchangeable, CCP aims to ensure valid coverage at the cluster level, i.e., \( P\bigl( y \in \widehat{C}(\mathbf{x}) \mid \kappa(\mathbf{x}) = k \bigr) \gtrsim 1 - \alpha \) for all \( k \in \{1,\dots,K\} \).

By incorporating structural information derived from drug and protein nonconformity scores patterns, this method offers more personalized and locally calibrated prediction intervals. The use of ECDF-based clustering allows to better capture heteroscedasticity and latent subpopulation effects, especially under challenging generalization scenarios such as drug and protein disjoint splits.

\subsubsection{Cluster-Conditioned Conformal Prediction: Feature-Based Clustering}
\label{sec:ccp_fc}
\begin{figure}[t]
	\centering % Often used to center the figure
	\includegraphics[width=0.75\linewidth]{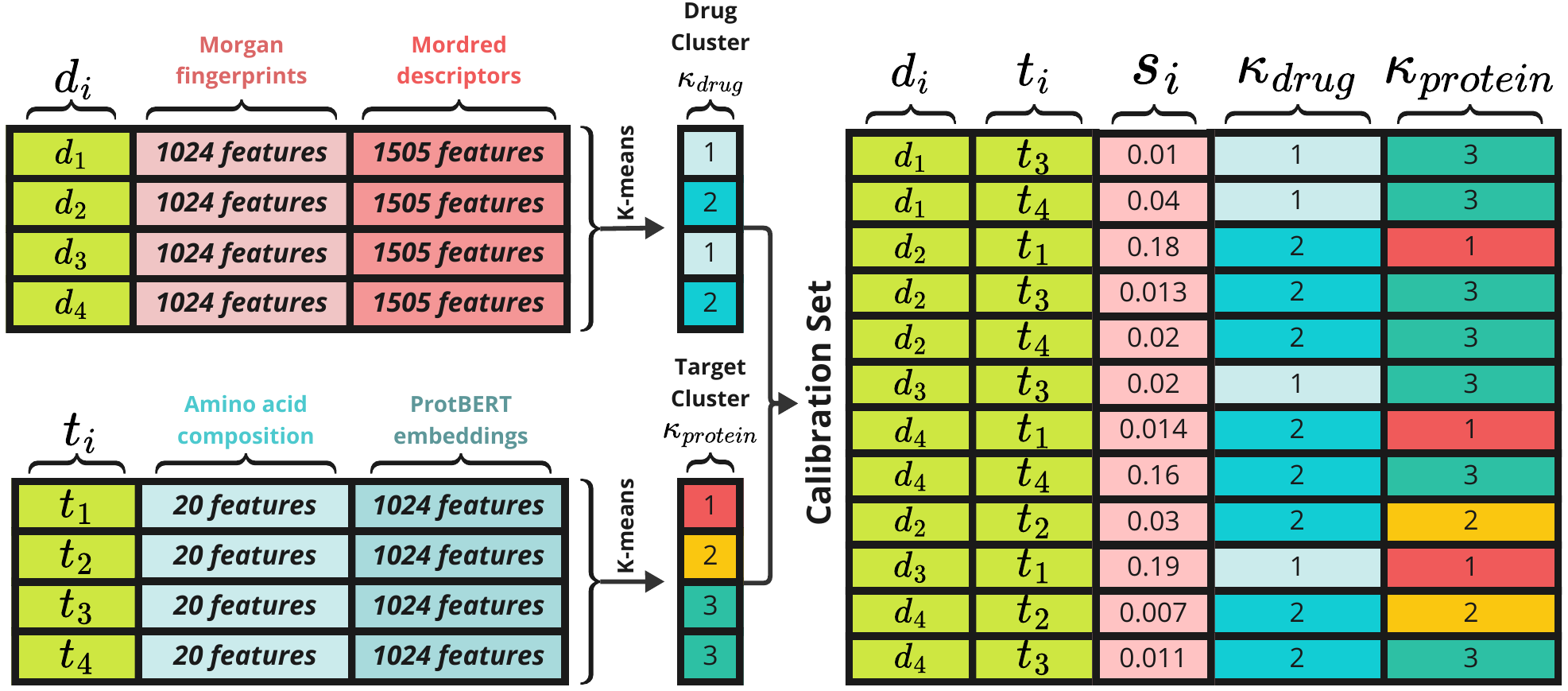}
	\caption{Two-level feature-based clustering in CCP-FC. Drugs \(d_i\) and proteins \(t_j\) are embedded using molecular and sequence features, respectively, and clustered via \(k\)-means to assign \(\kappa_{\text{drug}}\) and \(\kappa_{\text{protein}}\), used for grouping calibration scores and estimating cluster-specific quantiles.}
	\label{fig:feature_clustering} % Label should typically go after the caption
\end{figure}

In contrast to Nonconformity-Based Clustering, this variant, called CCP with Feature-Based Clustering (CCP-FC), groups calibration samples according to their original input feature representations. The goal is to leverage biologically or chemically meaningful similarities embedded in the feature space to enable more tailored uncertainty quantification. The feature vectors are the same as those described in Section~\ref{sec:dataset}, encompassing molecular fingerprints and descriptors for drugs, and sequence-based or embedding features for proteins.
Let \(\mathbf{x}_i\) denote the feature vector for the drug–target interaction \((d_i, t_j)\). We define a clustering function \(\kappa_{\mathrm{feat}} : \mathcal{X} \to \{1, \dots, K\}\) that assigns each calibration sample to one of \(K\) clusters based on similarity in feature space, using \(k\)-means clustering.
Following the same protocol as in Nonconformity-Based Clustering, we split the calibration set $\mathcal{D}_{\text{cal}}$ into two disjoint subsets: one used for clustering based on input feature vectors, and the other for computing cluster-level quantiles.
The split ratio is determined by a hyperparameter \(\gamma \in \{0.25, 0.5, 0.75\}\), which balances clustering fidelity and quantile estimation reliability.
Feature-based clustering is applied separately for drugs and proteins, as illustrated in Figure~\ref{fig:feature_clustering}: each drug \(d_i\) is represented by a concatenation of Morgan fingerprints and Mordred descriptors, then assigned to a cluster \(\kappa_{\text{drug}}(d_i)\) via \(k\)-means; each protein \(t_j\) is encoded using amino acid composition and ProtBERT embeddings, then assigned to a cluster \(\kappa_{\text{protein}}(t_j)\) through \(k\)-means.
Each calibration interaction \((d_i, t_j)\) is thus mapped to a cluster pair \((\kappa_{\text{drug}}(d_i), \kappa_{\text{protein}}(t_j))\). These indices are then used for quantile estimation. Cluster-specific quantiles are computed using Equations~\ref{eq:ccp_quantile_drug} and \ref{eq:ccp_quantile_prot}, the combined test-time threshold \(\hat{q}_{\mathrm{c}}\) is determined by Equation~\ref{eq:ccp_combined_quantile}, and the final prediction interval is constructed using Equation~\ref{eq:ccp_interval}.
To determine the best number of clusters \(n_{\text{clusters}}\) and the optimal \(\gamma\), we follow the same procedure as described in Section~\ref{sec:ccp_nc}.

\subsubsection{Cluster-Conditioned Conformal Prediction: Nearest Neighbor-Based Calibration}
\label{sec:ccp_nn}
This variant, referred to as CCP with Nearest Neighbor-Based Calibration (CCP-NN), eliminates the use of fixed global or cluster-based partitions by dynamically constructing a local calibration set for each test sample based on feature similarity \cite{orvsolic2023dynamic}. The main idea is to tailor the quantile estimation to the most relevant neighborhood of each test instance, allowing for finer-grained adaptation to local uncertainty patterns.
Let \(\mathbf{x}_i\) denote the feature vector corresponding to a test drug–target pair \((d_i, t_j)\). We define a dynamic local calibration set \(\mathcal{D}_{\text{cal}}(\mathbf{x}) \subset \mathcal{D}_{\text{cal}}\), constructed by identifying calibration interactions involving compounds and targets most similar to \(d_i\) and \(t_j\), respectively. Specifically, we compute the Tanimoto similarity \cite{rogers2010extended} between the test compound \(d_i\) and all compounds in the calibration set based on their Morgan fingerprints, as well as between the test protein \(t_j\) and all calibration proteins using binary feature profiles (e.g., amino acid composition or domain annotations). We then select the top 20 most similar neighbors for both \(d_i\) and \(t_j\). Finally, we extract all interaction pairs \((d_k, t_l)\) from \(\mathcal{D}_{\text{cal}}\) such that \(d_k\) belongs to the top 20 similar compounds and \(t_l\) belongs to the top 20 similar targets, yielding a neighborhood-specific subset of calibration interactions: \(\mathcal{D}_{\text{cal}}(\mathbf{x}) = \left\{ (\mathbf{x}_k, y_k) \in \mathcal{D}_{\text{cal}} \;\middle|\; d_k \in \mathrm{NN}_{20}^{\text{compound}}(d_i),\; t_k \in \mathrm{NN}_{20}^{\text{target}}(t_j) \right\}\).

From this dynamically constructed neighborhood \(\mathcal{D}_{\text{cal}}(\mathbf{x})\), we collect the associated nonconformity scores $s_k$ for each calibration sample interaction and sort them in non-decreasing order. The conformal quantile \(\hat{q}\) is then computed according to Equation~\ref{eq:quantile_calculation}, using the number of scores in the local set. The resulting prediction interval for \(\mathbf{x}\) is formed via Equation~\ref{eq:ccp_interval}.
This method offers highly adaptive and personalized calibration, particularly useful in heterogeneous or sparse interaction domains where global exchangeability assumptions are weak. By relying on similarity-based neighborhoods derived from domain-relevant metrics, it enables localized uncertainty quantification while still preserving conformal validity assumptions within the dynamic calibration subsets.

\subsection{Evaluation Metrics}\label{sec:evaluation_metrics}
CP provides a principled approach for constructing calibrated prediction intervals with finite-sample validity guarantees. To assess the performance of CP methods across different splitting and calibration set constructing strategies, we report several evaluation metrics that quantify both calibration (validity) and informativeness (efficiency).
The most fundamental validity metric is the empirical coverage probability, which measures the proportion of test samples whose true binding affinities fall within the predicted intervals:
\begin{equation}\label{eq:overall_coverage_dti}
    \text{Coverage}_{(1-\alpha)} = \frac{1}{n_{\text{test}}} \sum_{k=1}^{n_{\text{test}}} \mathbb{I}\bigl(y_k \in \widehat{C}_{(1-\alpha)}(\mathbf{x}_k)\bigr),
\end{equation}
where $\widehat{C}_{(1-\alpha)}(\mathbf{x}_k)$ denotes the prediction interval for sample $k$ at coverage level \(1 - \alpha\), and $\mathbb{I}\{\cdot\}$ is the indicator function. This metric reflects how well the empirical coverage aligns with the nominal target level.
To assess efficiency, we compute the average width of the prediction intervals:
\begin{equation}\label{eq:scp_width_dti}
    \text{Width}_{\text{avg}} = \frac{1}{n_{\text{test}}} \sum_{k=1}^{n_{\text{test}}} (\mathrm{upper}_k - \mathrm{lower}_k),
\end{equation}
where \(\mathrm{upper}_k\) and \(\mathrm{lower}_k\) are the endpoints of the interval for test instance $k$. Narrower intervals imply more confident and informative predictions, provided that validity is maintained.
To evaluate fairness and subgroup-wise calibration, we use the Mean Absolute Coverage Gap (MACG), which quantifies the deviation from the desired coverage within specific groups:
\begin{equation}\label{eq:macg_drug_full}
    \text{MACG}_{\text{subgroup}} = \frac{1}{D} \sum_{i=1}^D \left| \text{Coverage}_i - (1 - \alpha) \right|,
\end{equation}
where $D$ is the number of unique drugs, proteins, or clusters (depending on the subgroup type), and $\text{Coverage}_i$ is the empirical coverage for subgroup $i$, computed as:
\begin{equation}\label{eq:group_coverage_dti}
    \text{Coverage}_i = \frac{1}{|\mathcal{I}_i|} \sum_{k \in \mathcal{I}_i} \mathbb{I}\bigl(y_k \in \widehat{C}(\mathbf{x}_k)\bigr),
\end{equation}
with $\mathcal{I}_i$ being the index set for subgroup $i$ (e.g., all test samples involving drug $i$). We report the following metrics: MACG per Drug, which measures calibration gaps for individual compounds; MACG per Protein, which measures calibration gaps across different target proteins; and MACG per Cluster (applicable in Cluster-Conditioned Prediction), which evaluates how well CCP achieves localized calibration across learned drug and protein clusters.
In addition to numeric metrics, we also use reliability diagrams and coverage versus coverage plots to visualize empirical coverage across varying coverage levels \((1 - \alpha)\), providing further insight into the calibration quality of conformal predictors.

\section{Experiments and Results}
\label{sec:experiments}
This section presents results of evaluation of the CP methods introduced in Section~\ref{sec:cp_general}. We evaluated each method under the four dataset splitting strategies introduced in Section~\ref{sec:split_strategies}. All CP models were built upon the same predictive model architecture described in Section~\ref{sec:predModel}, but were trained separately using the training sets obtained from each splitting strategy. This ensured that the corresponding test sets remained completely unseen during model training. Below, we report the results of regression using the trained models on their respective test sets. Performance is measured using the Root Mean Squared Error (RMSE) and the coefficient of determination (\(R^2\)) for each splitting strategy. As shown in Table~\ref{tab:kiba_regression_results}, the predictive model achieves the best performance under the Random Split, while generalization becomes progressively more challenging in the Drug, Protein, and especially the New Drug–Protein Split, as reflected by increasing RMSE and decreasing \(R^2\) values.
\begin{table}[t]
    \centering
    \caption{Regression performance on the dataset using different data splitting strategies.}
    \label{tab:kiba_regression_results}
    \begin{tabular}{lcc}
        \toprule
        \textbf{Splitting Strategy} & \textbf{RMSE} & \textbf{\(R^2\)} \\
        \midrule
        Random Split           & 0.0207 & 0.5729 \\
        Drug-based Split       & 0.0225 & 0.3821 \\
        Protein-based Split    & 0.0256 & 0.4185 \\
        New Drug–Protein Split & 0.0266 & 0.1465 \\
        \bottomrule
    \end{tabular}
\end{table}

\begin{figure}[t]
	\centering % Often used to center the figure
	\includegraphics[width=0.65\linewidth]{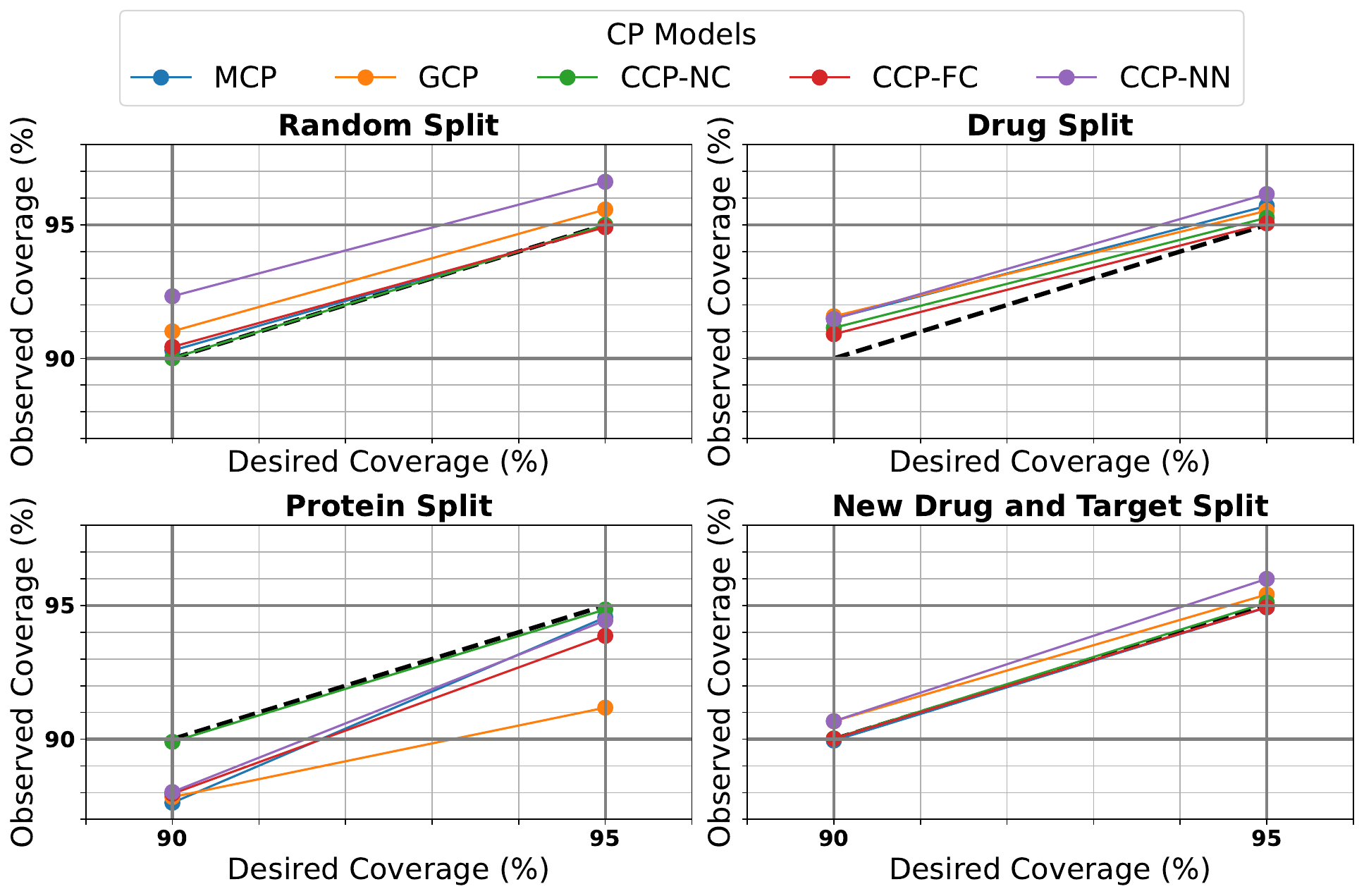}
	\caption{Observed vs.\ expected coverage across four dataset splitting strategies. Colors represent the six CP methods; the black diagonal indicates ideal coverage.}
	\label{fig:Confidence_vs_Observed} % Label should typically go after the caption
\end{figure}

We begin by assessing global coverage performance, as defined in Equation~\ref{eq:overall_coverage_dti}. Figure~\ref{fig:Confidence_vs_Observed} compares observed coverage against the expected coverage level for each CP method across the dataset splitting strategies. The observed deviations from the ideal coverage line may not necessarily indicate violations of the exchangeability assumption. Instead, they could partly be attributed to the use of a fixed test set shared across all CP methods. This shared test set might introduce consistent coverage deviations within each split, causing multiple methods to exhibit similar patterns of overcoverage or undercoverage.
In the Random Split, both CCP-FC and CCP-NC closely follow the ideal line, while MCP deviates slightly at the 90\% coverage. GCP and CCP-NN are unstable with observed coverages exceeding the nominal levels.
For the Drug Split, none of the methods achieve consistent coverage, but CCP-FC and CCP-NC are closest to the 95\% target. All methods are overconfident at 90\% and 95\%.  
In the Protein Split, CCP-NC achieves reliable coverage at both 90\% and 95\%. MCP and CCP-NN perform well at 95\%, whereas GCP and CCP-FC fall short.  
Under the New Drug--Protein Split, all methods yield reasonably accurate coverage, though GCP and CCP-NN show the weakest performance in this challenging scenario.

\begin{figure}[t]
	\centering % Often used to center the figure
	\includegraphics[width=0.6\linewidth]{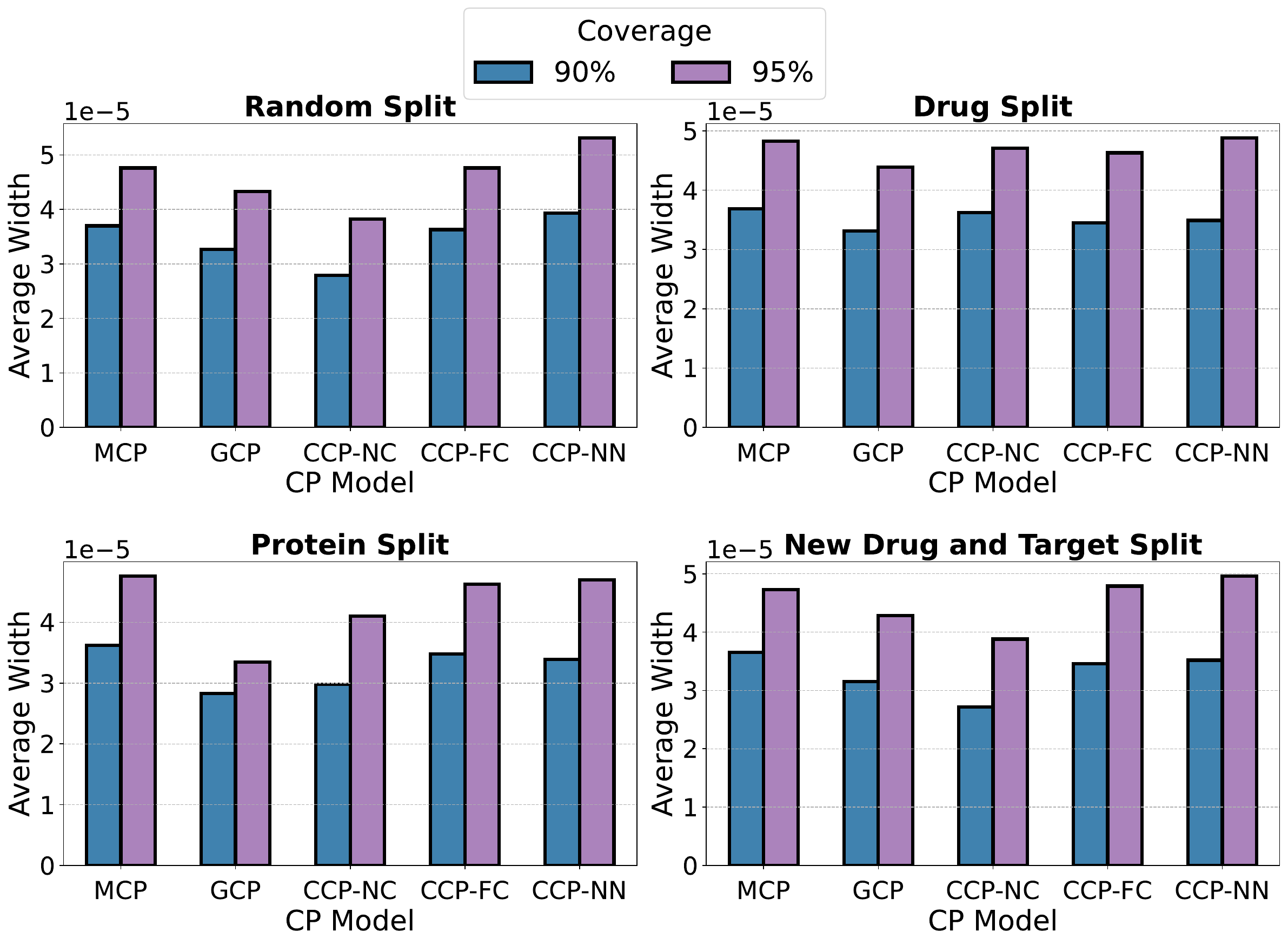}
	\caption{Mean prediction–interval width ($\times10^{-5}$) produced by five CP models—MCP, GCP, CCP-NC, CCP-FC, and CCP-NN—across four data–splitting strategies.}
	\label{fig:Average_Width} % Label should typically go after the caption
\end{figure}

We next assess the informativeness of CP methods by comparing their mean prediction interval widths (Figure~\ref{fig:Average_Width}). Narrower intervals indicate more confident and precise predictions, while wider intervals reflect higher uncertainty needed to maintain coverage guarantees.
CCP-NC produces the tightest intervals in both the Random and New Drug--Protein Splits at 90\% and 95\% coverage. In the Random Split, this sharpness stems from stable residual patterns across randomly partitioned calibration and test sets, resulting in lower nonconformity scores and tighter quantiles.
Prediction intervals tend to be narrower when model residuals are small—i.e., when \(f_\theta(\mathbf{x})\) closely approximates the true label \(y\). Additionally, in homoscedastic settings where the data noise is relatively uniform, nonconformity scores vary less across the input space, resulting in more efficient thresholds.
In the more challenging New Drug–Protein Split, CCP-NC maintains superior performance by leveraging its adaptive thresholding mechanism (Equation~\ref{eq:ccp_combined_quantile}) to condition on both drug and protein clusters. This strategy enables CCP-NC to select efficient quantiles even under data scarcity by pooling nonconformity scores from structurally similar clusters. 
By contrast, GCP attains narrower intervals in the Drug and Protein Splits through its group-based thresholding scheme (Equation~\ref{eq:gcp_combined_quantile}), which adapts intervals based on previously seen drugs or proteins. However, when both entities in a test pair are unseen—as in the New Drug–Protein Split—GCP becomes less reliable, whereas CCP-NC continues to perform robustly due to its cluster-based generalization.

\begin{figure}[t]
	\centering % Often used to center the figure
	\includegraphics[width=0.8\linewidth]{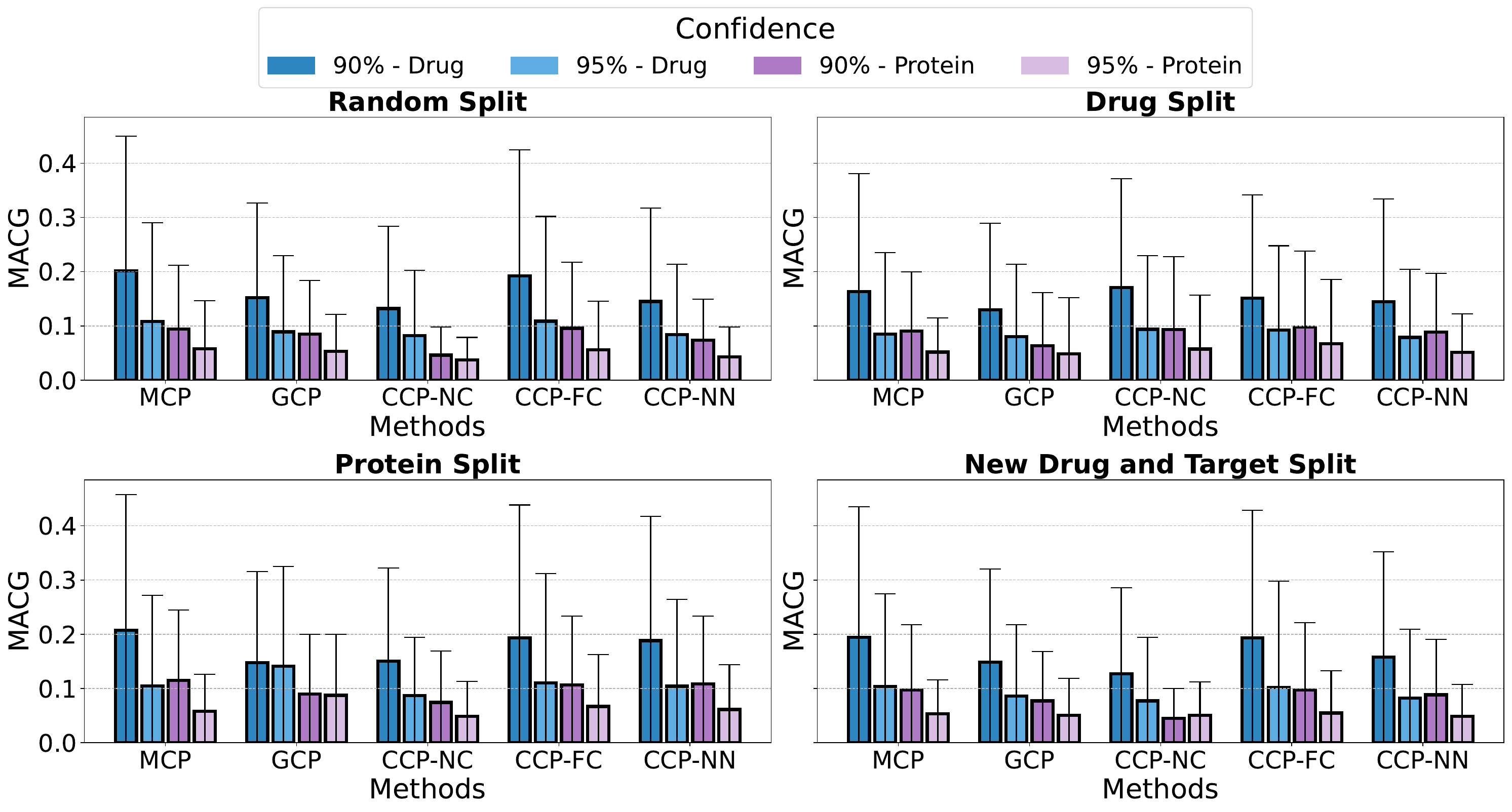}
	\caption{MACG (mean \(\pm\) STD) computed at the drug and protein levels in the test set. Within each split, bar height gives the MACG across all compounds or targets for five CP methods. Error bars represent standard deviations across subgroups.}
	\label{fig:mean_macg} % Label should typically go after the caption
\end{figure}

To assess subgroup-level coverage, we report MACG at the drug and protein levels (Figure~\ref{fig:mean_macg}). Each subplot corresponds to a data-splitting strategy, comparing five CP methods across coverage levels. Bars show mean MACG over test compounds or targets, with error bars denoting the STD of absolute coverage gaps. Lower values indicate smaller subgroup-level gaps and reduced uncertainty.
CCP-NC consistently achieves the lowest MACG in both the Random and New Drug–Protein Splits, for both drug and protein subgroups, demonstrating its robustness in both well-behaved and challenging generalization settings. In the Drug Split, GCP yields the lowest MACG at both the drug and protein levels. This reflects the strength of GCP’s entity-level conditioning: when a test drug is unseen but the associated protein is known, GCP computes quantiles using calibration interactions involving the same protein. Since there are only 229 unique proteins in the dataset—compared to 2068 drugs—each protein appears more frequently, resulting in more reliable quantile estimates. Consequently, GCP performs well when conditioning on the well-represented protein side of the interaction.
In contrast, CCP-NC outperforms other methods in the Protein Split, where the test proteins are unseen and the drugs are known. Due to the large number of unique drugs and their sparser individual representation, GCP’s conditioning on specific drugs becomes less stable. However, CCP-NC leverages clustering of nonconformity score distributions to group similar drugs and proteins, allowing it to pool calibration data across structurally or behaviorally similar entities. This generalization via cluster-level quantiles helps CCP-NC maintain low MACG even when facing data sparsity and unseen proteins.
Overall, while GCP performs well in settings where one entity is well-represented and directly observed in the calibration set (e.g., Drug Split), CCP-NC is more effective in scenarios where both entities may be sparse or unseen, as it generalizes via clustering. These complementary behaviors are evident in the Protein Split, where CCP-NC achieves the best MACG at both the drug and protein levels—except at the 90\% level for drugs, where its performance slightly trails.

\begin{figure}[t]
	\centering % Often used to center the figure
	\includegraphics[width=0.8\linewidth]{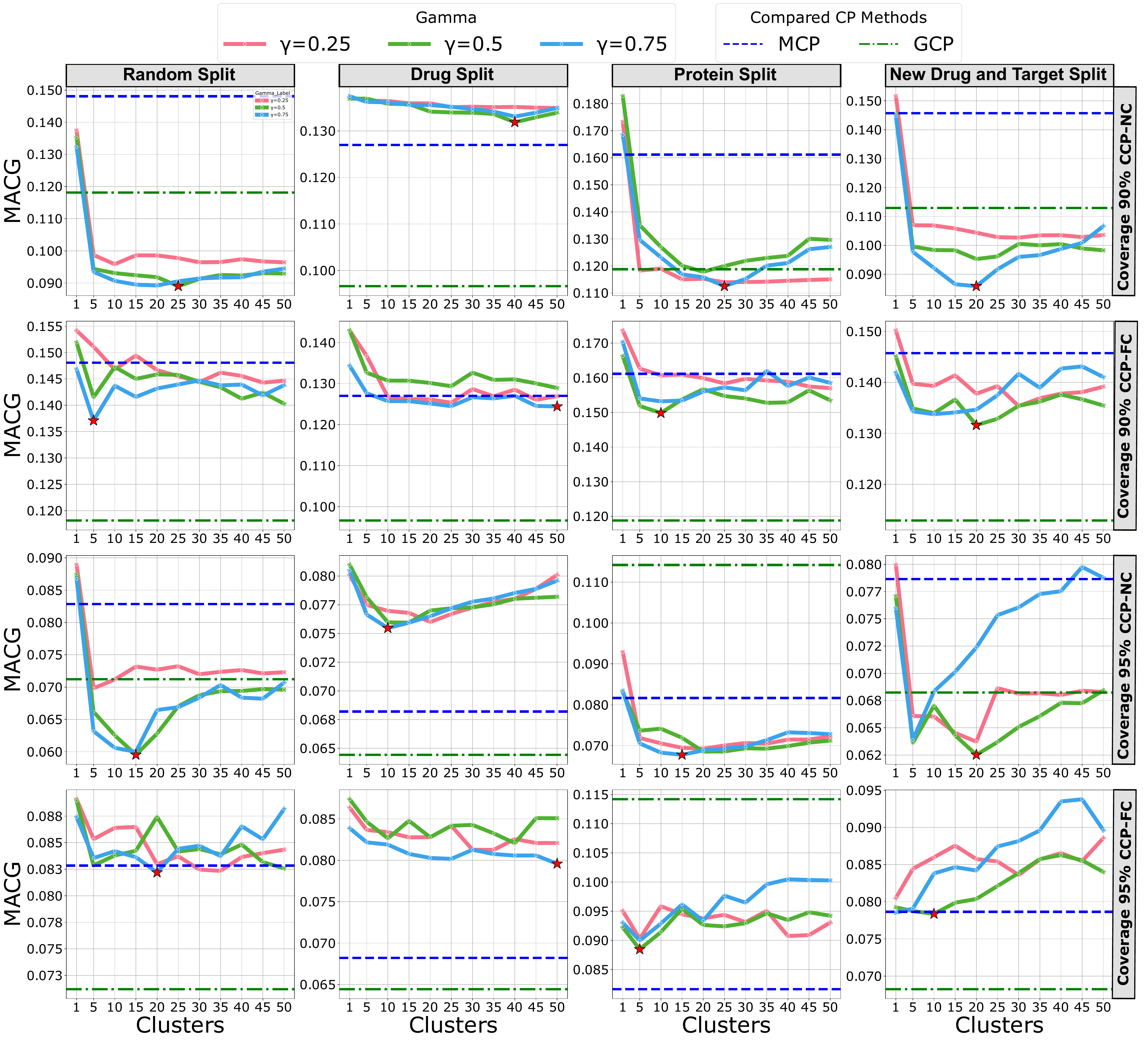}
	\caption{Combined MACG values for CCP-NC and CCP-FC across cluster counts, calibration set ratios ($\gamma$), desired coverage, and split strategies. Each subplot shows MACG variation with clustering parameters; colored curves indicate $\gamma$ values. Red stars mark the best configurations. Blue dashed and green dash-dotted lines show MCP and GCP baselines.}
	\label{fig:Combined_MACG_CCPNC_CCPFC} % Label should typically go after the caption
\end{figure}

The results presented above for CCP-NC and CCP-FC correspond to their best-performing configurations, selected from a grid of 33 combinations of calibration set split ratios \(\gamma \in \{0.25, 0.50, 0.75\}\) and cluster counts \(n_{\text{clusters}} \in \{1, 5, 10, 15, 20, 25, 30, 35, 40, 45, 50\}\). For each setting, we evaluated performance using the combined MACG across both drug and protein subgroups, as defined in Equation~\ref{eq:combine_MACG}, and selected the configuration that minimized this value.
Figure~\ref{fig:Combined_MACG_CCPNC_CCPFC} illustrates the combined MACG values achieved by CCP-NC and CCP-FC across all tested parameter combinations, covering different coverage levels and dataset splitting strategies. Each subplot corresponds to a specific pair of coverage level and data split. Colored lines represent different calibration ratios \(\gamma\), while the horizontal axis denotes the number of clusters. Red stars highlight the optimal configuration that produced the lowest combined MACG. For reference, baseline performance is indicated by two horizontal lines: a blue dashed line for MCP and a green dash-dotted line for GCP.

\section{Conclusion}\label{sec:conclusion}
Based on the obtained results, it is evident that the choice of dataset splitting strategy has a substantial impact on the performance of CP methods. Among all methods, CCP-NC demonstrates better performance in terms of uncertainty quantification, consistently producing tighter prediction intervals and achieving lower subgroup-level coverage gaps (MACG) in both the Random Split and the most challenging New Drug–Protein Split. This highlights its robustness under both favorable and generalization-heavy settings.
In the Drug Split, GCP outperforms other methods in terms of interval width, benefiting from its ability to condition directly on well-represented proteins. In the Protein Split, GCP yields narrower intervals overall, yet its subgroup-level calibration is less stable: CCP-NC performs better at the 90\% coverage level, whereas GCP performs slightly better at 95\%.
Our findings suggest that clustering-based conformal predictors—particularly those leveraging nonconformity score distributions like CCP-NC, offer a better solution for reducing uncertainty in drug–target interaction datasets. By capturing structure across both drugs and proteins, CCP-NC provides a data-efficient approach to generating tighter prediction intervals. These insights may extend to other interaction-centric domains, where samples depend on two or more structured entities, making residual-driven clustering a powerful tool for uncertainty quantification.

\bibliographystyle{unsrt}  
\bibliography{references}  %%% Remove comment to use the external .bib file (using bibtex).
\end{document}